\documentclass{article}
\usepackage{acra}
\usepackage{amsmath, amssymb}
\usepackage{graphicx}
\usepackage[utf8]{inputenc}
\usepackage{amsmath}
\usepackage{resizegather}
\usepackage{url}

\title{Visual based Tomato Size Measurement System for an Indoor Farming Environment}

\author{Andy Kweon, Vishnu Hu, Jong Yoon Lim, Trevor Gee, Edmond Liu, \\ \textbf{Henry Williams, Bruce A. MacDonald, Mahla Nejati, Inkyu Sa, and Ho Seok Ahn*} \\ 
Centre for Automation and Robotic Engineering Science\\ 
The University of Auckland, New Zealand\\
\{skwe902, vhu474, eliu587\}@aucklanduni.ac.nz, \\ \{jy.lim, t.gee, henry.williams, b.macdonald, m.nejati, hs.ahn\}@auckland.ac.nz, enddl22@gmail.com }

\begin{document}

\maketitle

\begin{abstract}
        
As technology progresses, smart automated systems will serve an increasingly important role in the agricultural industry. Current existing vision systems for yield estimation face difficulties in occlusion and scalability as they utilize a camera system that is large and expensive, which are unsuitable for orchard environments. To overcome these problems, this paper presents a size measurement method combining a machine learning model and depth images captured from three low cost RGBD cameras to detect and measure the height and width of tomatoes. The performance of the presented system is evaluated on a lab environment with real tomato fruits and fake leaves to simulate occlusion in the real farm environment. To improve accuracy by addressing fruit occlusion, our three-camera system was able to achieve a height measurement accuracy of 0.9114 and a width accuracy of 0.9443.

\end{abstract}

\section{Introduction}
Due to rapid urbanization, population growth, and increasing wealth in developing countries, an increase of approximately 25–70\% above the current production levels may be sufficient to meet crop demand in 2050 \cite{Hunter2017}. To meet this need, precision agriculture techniques have gained attraction, as it also meets the economic and environmental pressures of arable agriculture \cite{Stafford2000}. 

Precision agriculture is the application of technologies and ideas to manage all temporal and spatial variables associated with farming, to maximize crop yield while reducing environmental strain. For precision agriculture, assessing, managing, and evaluating crop production in the space-time continuum \cite{PierceNowak1999} is crucial for success. To achieve this, intelligent systems based on computer vision algorithms to identify plant's physical and physiological
traits \cite{Chandra2020} are being researched by the scientific community, which speeds up the plant monitoring process compared to manual observation. 

We have been developing an indoor smart farming system for monitoring fruit growth utilizing deep neural networks and a cost-effective camera system. We detect targets, i.e. tomato and leaves, and count the number of branches regularly, i.e. every 3 days. We also measure how ripe tomato is and estimate the timing to harvest. We record their growing on our server system and visualize the captured images on our client system. 

In this paper, We focus on how to find tomato and measure its size. We implement affordable camera system using Intel Realsense camera \footnote{\url{https://www.intelrealsense.com}}, which costs around USD 300 per camera, instead of using high performance camera, i.e. Zivid camera \footnote{\url{https://www.zivid.com/}} that costs around USD 7000 in this research. We propose a novel approach to measure the target size using cheaper camera. From our experiment, we confirm our approach gives reasonably accurate results compare to expensive cameras. 


\section{Related Works}

\subsection{Camera System}
For 3D reconstruction with multiple frames, extrinsic camera calibration needs to be performed. One popular implementation requires a computer detectable marker such as a checkerboard or ArUco-board be scanned across cameras. Then, the transformation between the estimated board pose can be computed using an inverse matrix transform. This requires the front plane of the board to be in view of all three cameras. This strategy can be extended to 3D markers for cameras outside the board's field of view [Chai et al., 2018].

One similar paper featured 3D reconstruction across multiple frames using a single RealSense D435i camera [Masuda, 2021]. The work used depth information to reconstruct a point cloud, where semantic segmentation was applied to estimate leaf area. Notably, the paper did not use alignment by camera extrinsic calibration, but performed alignment by local feature mapping instead. A significant limitation that they listed was occlusion, as the stem of the plant needed to be seen for their calculations.

Various other research have investigated fruit identification and crop load estimation using different cameras, such as spectral camera \cite{Safren2007} \cite{Kim2004}, infrared camera \cite{Wachs2007} and time of flight (TOF) camera \cite{Vazquez2016}. However, these researches have found that while these cameras are superior in image quality for fruit identification, they were very difficult to operate and very expensive.

\begin{figure*}[!htbp]
      \centering
            {\includegraphics[width=5in, height=7cm]{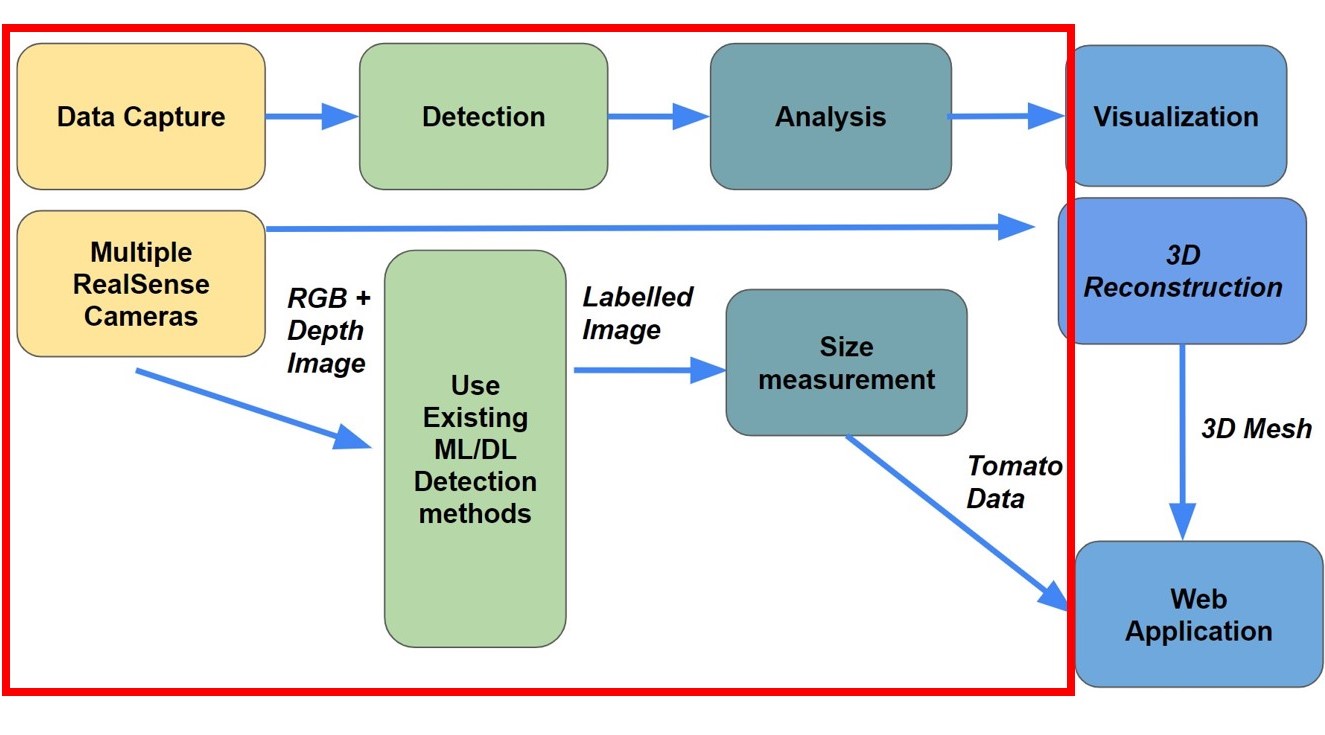}}
      \caption{System Diagram of our proposed plant monitoring system. In this paper, we focus on data capturing, detection, and size measurement as a part of analysis, outlined inside the red box, but not on visualization.}
      \label{fig:systemdiagram}
\end{figure*}

\subsection{Fruit Detection}
Different segmentation/fruit identification methods have been identified, ranging from simple threshold-based algorithms to deep learning implementations. Much research has been done on identifying fruits using RGB images, as it allows the researchers to analyze the image using the color, texture, and geometry of the image. Before deep learning became popular, researchers used traditional image processing techniques to segment fruits from the background. 

Threshold-based method (Otsu’s method) was used by \cite{Fu2019} to detect kiwifruits. RGB kiwifruit image was converted into gray-scale, and the grey image pixels were divided into fruit and background through thresholding. Through this method, an accuracy of 92\% at nighttime with flash was achieved. This method has shown that it is simple and efficient, but the limitations are that the results depend on the gray-scale value and have no spatial details. 

Edge detection methods, mixed with a circular fitting algorithm by \cite{Patel2012} were able to fit an approximately circular shape of the fruit, resulting in an efficiency of 98\% and an average yield measurement error of 31.4\%. However, occluded, and clustered fruits caused the algorithm to detect wrong number of fruits in an image. 

Color-based methods, using the idea that the color of the fruit is usually different from the background have also shown to be effective in segmenting fruits. \cite{Thendral2014} have compared the results of edge-based and color-based algorithms and found that color segmentation outperforms edge-based segmentation in detecting oranges with an accuracy of 85\%. The limitation however is that it requires the fruit to have a color difference from its surroundings. 

Region-based methods, such as watershed have been used to segment fruits as well, resulting in an accuracy of 98\% by \cite{Zeng2009} and a mean error of 3.091\% by \cite{Deepa2011}. However, researchers have noted that complex algorithm, heavy computing, and images requiring good contrast between regions are the limitations of the method.

In recent years, deep learning based methods of fruit detection have been popular, with some researchers using a DCNN model to detect tomato plant organs. A deep neural network based on VGGNet was used with 8-layer structure, and the researchers were able to achieve an average precision of 81.64\% for detecting tomato fruits \cite{Zhou2017}. Another group of researchers used a Mask R-CNN model to detect tomatoes for fruit counting from images taken from Intel Realsense D415 cameras. The researchers found the Mask R-CNN model implicitly learned object depth, which was necessary for separating the background elements from the foreground fruits. Furthermore, the researchers found that the Mask R-CNN model exceeded the performance of previous works, with the recall value exceeding the prediction accuracy of 0.91 \cite{Afonso2020}. However, occlusion stands as a major problem for image based semantic segmentation, with the best performing deep learning methods achieving around 90\% precision score for yield estimation \cite{Koirala2019}. Other factors such as lighting also influenced the overall accuracy, with some models generalizing better. 

\subsection{Fruit Size Measurement}
Many previous research have been done to measure fruit size using various techniques. One research developed an apple sizing algorithm based on pixel size. The relationship between pixel size, and distance from the camera was modeled using checkerboard images placed at ten different distances from the cameras. A regression model was made to predict pixel size based on pixel coordinates and the distance from the camera. Using this relationship, the major axis of each apple was measured by summing the physical size of all the pixels along the major axis, resulting in a mean absolute error of 15.2\% and accuracy of 84.8\% \cite{Gongal2018}. 

Some other researchers used Intel RealSense D415 cameras to measure different vegetable sizes and using keypoints detection to find top, bottom, left and right points of the fruit. From these keypoints and the depth reading from the RealSense camera, the researchers mapped the 2D pixel distance to 3D real-world distance, resulting in a MAPE of 1.99\% for diameter and 3.90\% for the length of tomato fruits when the fruits were 40cm away from the camera \cite{Zheng2022}. However, both of these methods were limited in real-world use as they were both tested on fruits with an unoccluded view. 

Another group of researchers developed a tomato volume estimating robot using a motion stereo camera, and tried to measure tomato sizes on a fake plant bed to simulate occlusion. To estimate the fruit width, the researchers combined a stereo image to create a 3D point cloud, and k-means method is used to extract the fruit region as 2D and 3D from the point cloud. The 2D region is used to fit an outline circle, to estimate the occluded areas of the fruit. From this outline, an estimate of the fruit width can be calculated, by converting the 2D pixel distance to 3D distance. This width estimation method gave a mean absolute percentage error of around 5\% from the selected data. However, the researchers used saliency map based vision processing method, which was sometimes inaccurate in estimating the tomato outline, causing the MAPE error to be 40\% in some fruit measurements \cite{Fukui2018}.
 
\section{Tomato Growing Monitoring System}

\subsection{System Overview}
Our monitoring system involves four steps, data capture using multiple Realsense cameras, deep learning detection model to find ripeness of the fruit, the bounding box and segmentation of individual tomatoes from the captured RGB image and a size measurement algorithm to estimate tomato fruit width and height and 3D reconstruction to display the 3D mesh in the web application. The system diagram is shown in Figure \ref{fig:systemdiagram}, and in this paper we will discuss more about the data capture, detection and analysis methods of our system.

\subsection{Camera System and Dataset}
\begin{figure}[!htbp]
      \centering
            {\includegraphics[width=3.5in, height=6cm]{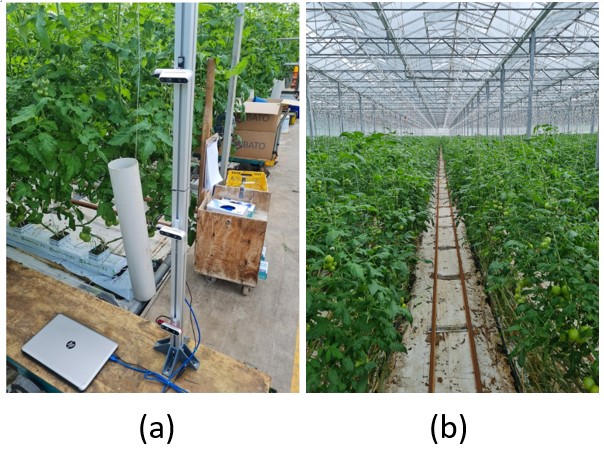}}
      \caption{(a) our camera system using three Intel Realsense cameras, (b) tomato farm environment where we do data capturing and field test.}
      \label{fig:farm}
 \end{figure}

Our capture system uses three Realsense D435 cameras, approximately placed at heights 15, 60, and 105 cm apart. The cameras are mounted on a vertical stand, 45 degrees apart, facing towards the plants roughly 60 cm away. The vertical stand was then mounted onto a cart, which would be pushed along the rails as shown in Figure \ref{fig:farm}(b). The three cameras were then connected to a laptop using a USB 3.0 hub and all camera intrinsics were calibrated in the same environment using Intel Realsense self camera calibration tool. 

The Realsense cameras are configured to take 1280 x 720p color image, along with 640 x 480p depth image, in order to reduce the bandwidth on the USB 3.0 bus. To match the resolutions, the depth image was then aligned with the color image, resulting in 1280 x 720p color and depth images.  The reconstructed point cloud generated from Intels in-built ROS drivers was also saved, along with the camera extrinsics. Sampling was done roughly taken at the same time of day, at 11am in both cloudy and clear environments. Samples were taken on three winter dates, from mid July to early August. 

From there, samples were taken from 25cm increments marked on the rails between tomato rows. A total of 140 samples were taken from different days, totalling 420 images from different frames.

The dataset was sampled from NZ-hothouse, Auckland, New Zealand. The location features a tomato greenhouse where the humidity, temperature and wind were all controlled. The first row sampled is a young Merlinice tomato plant roughly 10 weeks old, while the second and third rows are older Merlice tomato plants 10 months old. 

\subsection{Extrinsic Calibration}
\begin{figure}[!htbp]
      \centering
            {\includegraphics[width=2.3in, height=1.3in, angle=0,origin=c]{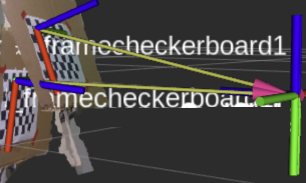}}
      \caption{Checkerboard markers as seen by the camera, which is used for calibration.}
      \label{fig:checkerboard}
 \end{figure}

To calibrate the camera extrinsic, a laminated A3 sized checkerboard, as shown in figure \ref{fig:checkerboard}, was held out between the cameras. The pose of the checkerboard was calculated using ros tuw package, which an opencv-based implementation to calculate the relative pose of the checkerboard to the camera.

3D pose and transformation can be expressed as a 4x4 homogeneous matrix in the form
\begin{equation}
    \begin{bmatrix}
                R_{11} & R_{12} & R_{13} & T_x\\
                R_{21} & R_{22} & R_{23} & T_y\\
                R_{31} & R_{32} & R_{33} & T_z\\
                0 & 0 & 0 & 1\\
    \end{bmatrix}
\end{equation}
with a 3x3 rotation matrix r and 3x1 translation matrix t. Since both poses represent the same pose in the world coordinate, a transformation matrix T needs to be applied to one of the poses to calibrate the frames. Consider the expression
\begin{equation}
P_{mid} = T \times P_{top} ,
\end{equation}
where T is the transformation between the coordinate frames. We substitute the relative pose of checkerboard as P, then solve the equation for transformation matrix T. Each transformation is saved then applied to align the coordinate frames during analysis.

\subsection{Lab Testing Environment}
\begin{figure}[!htbp]
      \centering
        \parbox{3in}
            {\includegraphics[width=3in, height=9cm]{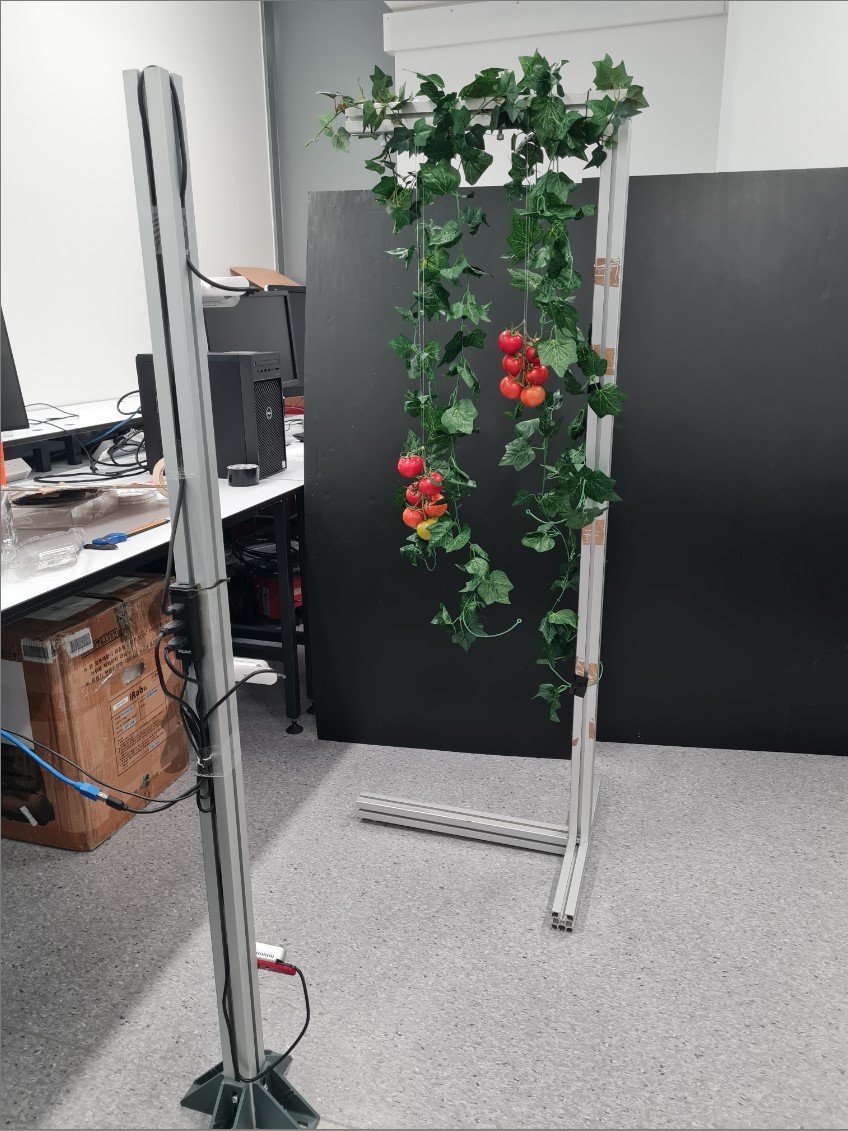}}
      \caption{Our lab testing environment with two real tomato trusses and artificial leaves}
      \label{fig:labtest}
 \end{figure}

To validate the effectiveness of our size measurement algorithm, a testing environment was set up to simulate the farm environment outlined in section 3.2. Two real tomato trusses containing 6 fruits were used to simulate the orchard environment. Based on our visits to the farm and through discussion with the farmers, the fruits of interest were 70-80cm off the ground (which we called the “Goldilocks zone”) and the trusses were around 10-20cm apart from each other. Based on these requirements, we have set up the testing environment as shown on \ref{fig:labtest}. Fake leaves were used to simulate occlusion like the orchard environment. 

\section{Tomato Fruit Detection}
\begin{figure}[!htbp]
      \centering
            {\includegraphics[width=3.1in, height=6cm]{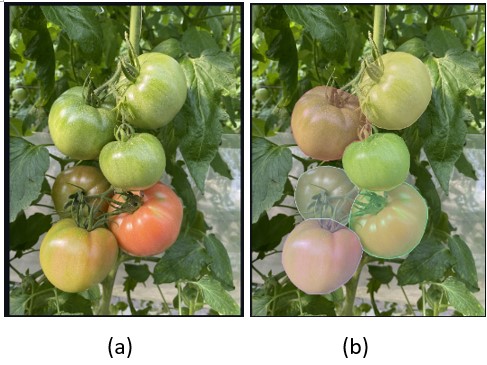}}
      \caption{(a) Sample image from the LaboroTomato dataset, (b) Segmentation mask of the LaboroTomato dataset.}
      \label{fig:Laboro}
\end{figure}

For our tomato detection, we have used a pretrained model from LaboroTomato big dataset, which contains 353 training images of tomatoes and 89 testing images. The tomatoes were labelled as fully ripened (greater than 90\% red colour), half ripened (30-89\% red colour) and green (less than 30\% red colour). Figure \ref{fig:Laboro} shows the sample image and the segmentation mask from the LaboroTomato dataset, which the model was trained on. 

\begin{figure}[!htbp]
      \centering
        \parbox{3in}
            {\includegraphics[width=3in, height=9cm]{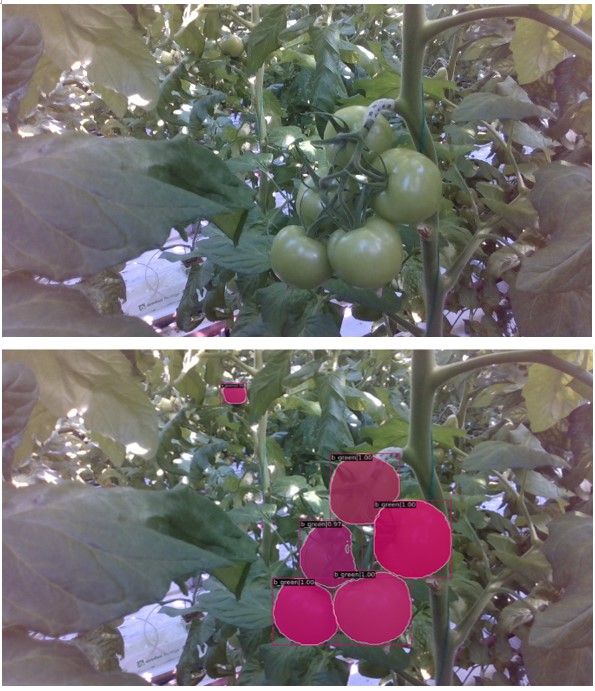}}
      \caption{Captured image of tomatoes from the farm using D435 camera (top), and detection result from Mask R-CNN pretrained model (bottom).}
      \label{fig:farmImage}
\end{figure}

Based on the testing dataset, the model has shown bbox AP of 67.9 and a mask AP of 68.4. The detection result on a captured image from a farm can be shown on Figure \ref{fig:farmImage}. The Mask R-CNN model is an extension of Faster R-CNN \cite{He2017}, which adds a branch for predicting segmentation mask on each Region of Interest in parallel with the existing branch for classification and bounding box regression.

From LaboroTomato, the pretrained Mask R-CNN model has a R-50-FPN 1x backbone, with training parameters of learning rate = 0.01, step = [32, 44] and 48 epoch. With the segmentation and the bounding box from the Mask R-CNN model, we used these information in conjunction with the depth map captured from D435 camera to try measure the height and width of tomatoes.

\section{Tomato Fruit Size Measurement}
To aid farmers in harvesting, it is important to accurately estimate size of tomatoes for yield estimation. With the captured RGB images and their depth maps from our three camera system, we have investigated a method to accurately measure tomato height and size from the detection result of the Mask R-CNN model. In this section, we outline the method of calculating fruit size, using the three camera system.

\subsection{Calculating Distance from 3D Co-ordinates}
The idea of calculating distance from 3D coordinates is simple, which is taking the square root of the sum of the squares of the differences between corresponding coordinates to calculate the distance in xyz space. The height of a tomato is simply the distance between the topmost point to the bottom-most point and the width of a tomato is between the leftmost and the rightmost point. 

\begin{figure}[!htbp]
      \centering
        \parbox{3in}
            {\includegraphics[width=2.8in, height=7cm]{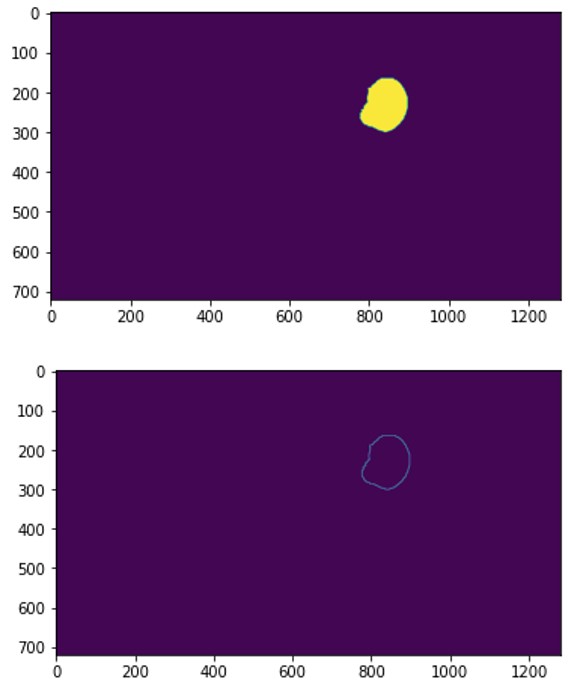}}
      \caption{Binary mask of tomato segmentation (top), and Binary mask after edge detection (bottom).}
      \label{fig:segmentation}
\end{figure}

\begin{figure*}[!htbp]
      \centering
            {\includegraphics[width=4.5in, height=8cm]{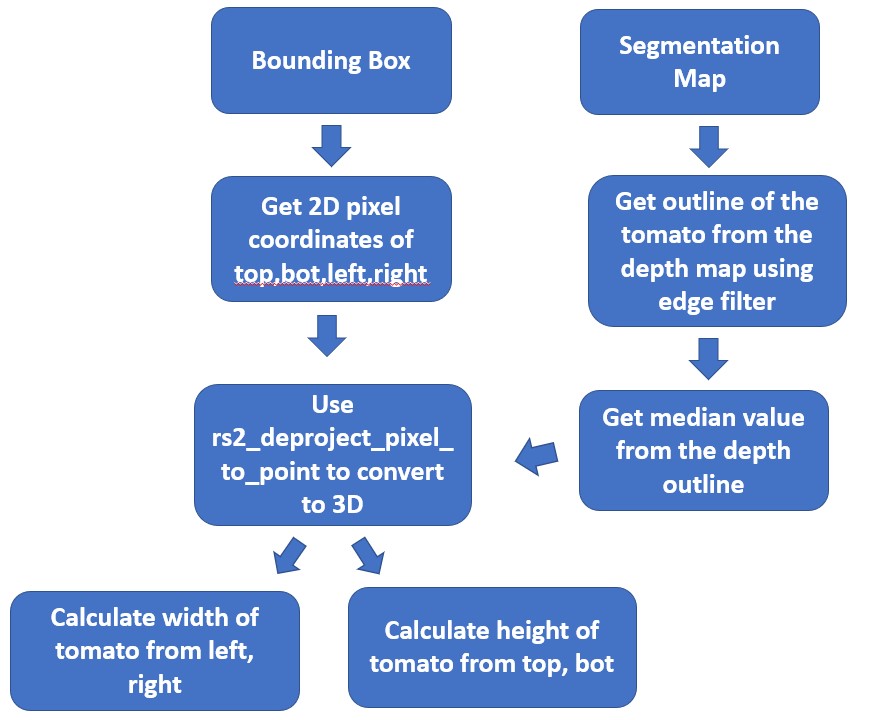}}
      \caption{Flowchart of tomato measurement method.}
      \label{fig:flowchart}
\end{figure*}

From running our pretrained Mask R-CNN model on the RGB image captured from the D435 camera, we can extract the pixel location of tomato bounding box (segmentation) from our RGB image. Using the corresponding depth image, we can use the same pixel coordinates of the bounding box (segmentation) to extract the depth data of the pixel. With this depth information and the pixel coordinates, Intel Realsense offers a function rs\_deproject\_pixel\_to\_point which returns a 3D x,y,z real world coordinate by inputting the pixel x,y location and the depth information. By using this function, we were able to acquire the x,y,z real world coordinates of top, bottom, left and right points of a tomato and they were used to calculate the distance between the two points using the formulas outlined in Eq. (3) and Eq. (4). 
    \begin{equation}
        \resizebox{.9\hsize}{!}
        {${Height} = \sqrt{{top_x - bot_x}^2 + {top_y - bot_y}^2 + {top_z - bot_z}^2},$}
    \end{equation}
    \begin{equation}
        \resizebox{.9\hsize}{!}
            {${Width} = \sqrt{{left_x - right_x}^2 + {left_y - right_y}^2 + {left_z - right_z}^2}.$}
    \end{equation}

This method is theoretically more accurate than mapping pixel distance to 3D space method but in practical applications, acquiring depth information around edges of a fruit is difficult \cite{Zheng2022}, resulting in a less accurate measurement. To overcome this and get more accurate edge depth, one assumption we made was that in a tomato fruit, the outermost points of a tomato (top, left, bottom, right) would have the same depth as each other. To estimate this depth from the segmentation map of a tomato, we acquired only the edge pixels of the tomato using an edge detector. From each individual tomato binary mask, a 3x3 matrix 
\begin{equation}
\begin{bmatrix}
            -1 & -1 & -1\\
            -1 & 8 & -1\\
            -1 & -1 & -1\\
\end{bmatrix}
\end{equation}
was applied to give a binary mask of the edges shown in Figure \ref{fig:segmentation}. 

Using these edge pixel locations, we got the corresponding depth values from the depth map and calculated the median depth value to remove any outlier values. The flowchart outlining the process of the tomato measurement method can be seen on Figure \ref{fig:flowchart}.

\subsection{Fill Ratio Estimation}
\begin{figure}[!htbp]
      \centering
        \parbox{3in}
            {\includegraphics[width=3in, height=14cm]{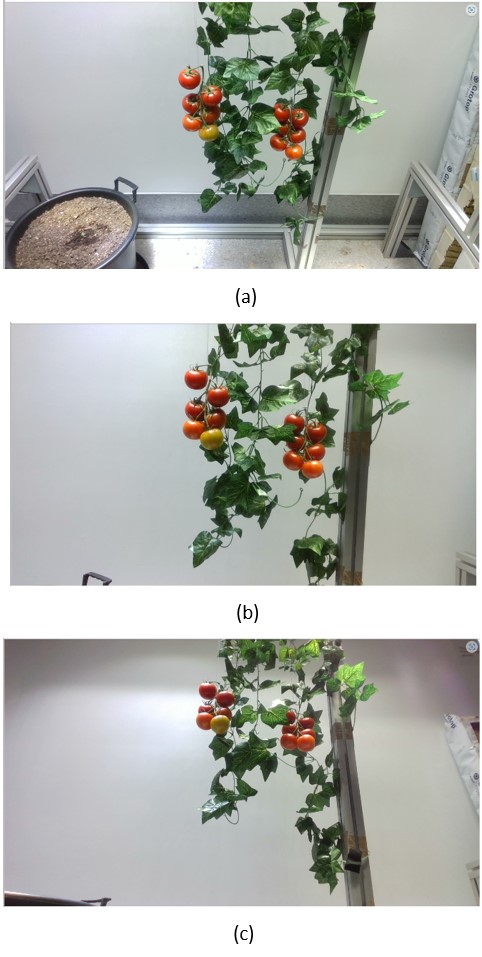}}
      \caption{Captured RGB images from our camera system using 3 Realsense cameras. (a) Top camera, (b) Middle camera, (c) Bottom camera.}
      \label{fig:cameras}
\end{figure}

Our 3 camera implementation allows tomato fruit images to be captured in 3 different angles. Through different angles, the system could be more robust against occlusion compared to having just a single camera pointing at the plant. Figure \ref{fig:cameras} shows the images captured from all three cameras, showing the different angles the system is able to capture. 

\begin{figure}[!htbp]
      \centering
        \parbox{3in}
            {\includegraphics[width=3in, height=4.5cm]{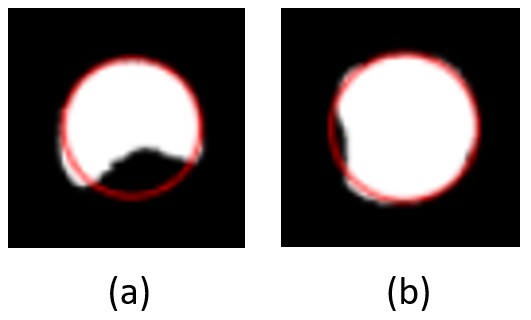}}
      \caption{Binary mask of a tomato fruit and the fitted circle (in red). (a) Fill ratio of 0.73, (b) Fill ratio of 0.94.}
      \label{fig:binarymask}
\end{figure}

Since our measurement method measures the distance between the visible points of the fruit, we assumed that images of fruits with less occlusion would give us a more accurate result. Using the circle fitting method implemented in \cite{Fukui2018}, an estimated circle was fit on the binary masks of tomato fruits shown on Figure \ref{fig:binarymask}. Based on the fitted circle, a fill ratio was calculated by Eq. (6), which gives an estimation of the amount of occlusion for each fruit. The higher the fill ratio, less the amount of occlusion on a fruit. 
\begin{equation}
        \text{Fill ratio} = \frac{\text{No. of tomato mask pixels inside fitted circle}}{\text{Pixel area of fitted circle}}.
\end{equation}

Based on this fill ratio, the measurement with the highest fill ratio was selected for each fruit. 

\subsection{Coordinate detection}
\begin{figure}[!htbp]
      \centering
        {\includegraphics[width=3.5in, height=10cm]{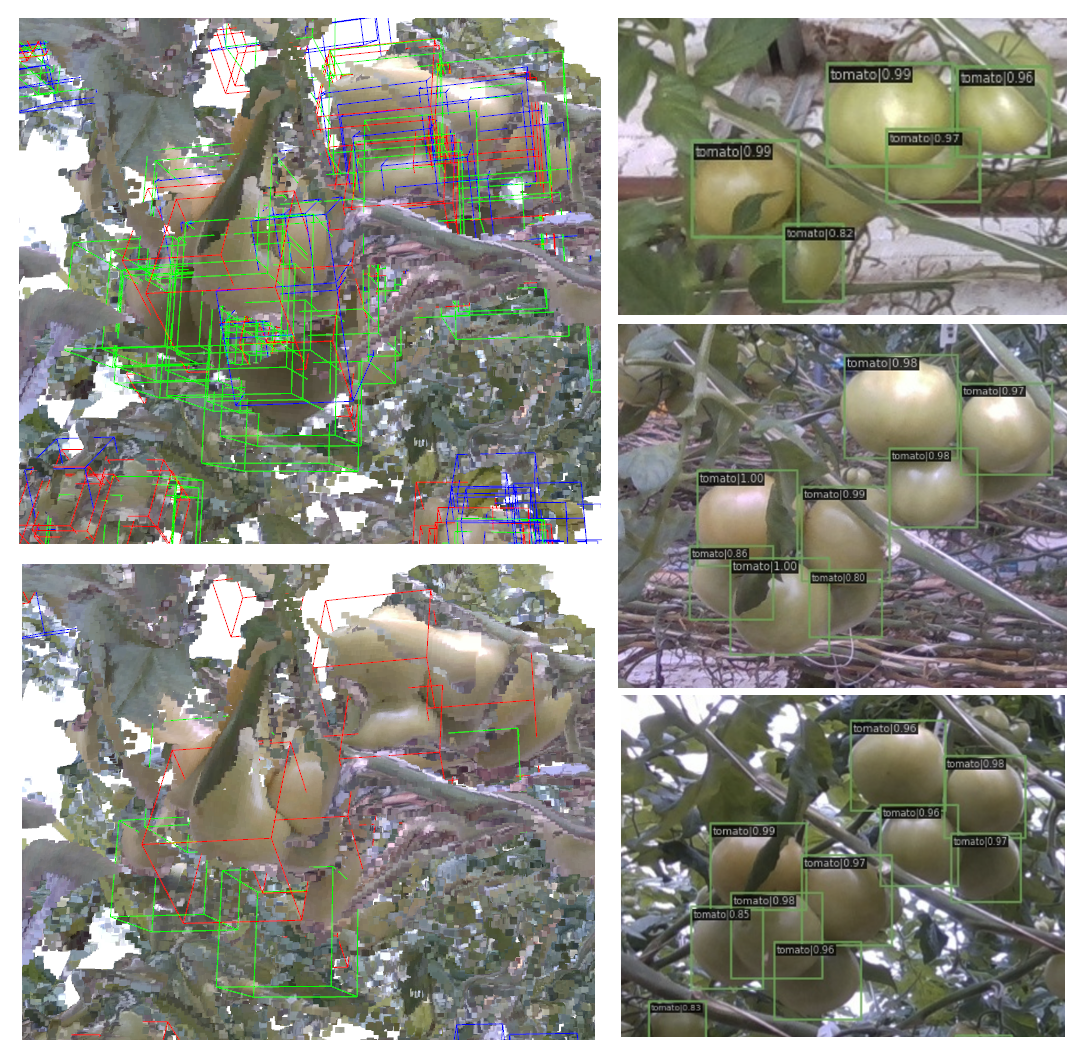}}
      \caption{Experimental results. Frequency map of detection (top-left), Detected fruits (bottom-left). 2D detection results from the nearest top, middle, and bottom cameras shown on the right.}
      \label{fig:results}
\end{figure}

To detect the location of the fruit relative to the world frame, the relative coordinate was first found using the Realsense cameras in-built stereo module. We used a projected 3D coordinate of the center of the bounding box, adding its estimated radius to its distance. Then the coordinate was transformed to the world frame using the transformation between the camera extrinsics as defined in 3.3. Due to many overlapping samples, the an average tomato is detected 6 times by different camera frames. We used a simple geometric-based location matching to identify overlapping detections. Fruits detected within the radius of each other were considered the same fruit. Figure \ref{fig:results} shows the results of the detection. A limitation of this method is that it is reliant on camera calibration, and depth data. We found the real-sense camera to have high depth error, which significantly increased error for fruits near the edge of the camera frame.

\section{Experimental Results and Discussion}
Using the 12 tomato fruits on the lab testing environment outlined in section 3.3, we calculated the RMSE error using Eq. (7) and accuracy using Eq. (8) of both height and width measurements of tomatoes. The mean tomato height was 39.4mm and the mean tomato width was 46.7mm for the 12 tomatoes in the lab test. 

\begin{equation}
 RMSE = \sqrt{\frac{1}{n}\Sigma_{i=1}^{n}{\Big(\frac{d_i -f_i}{\sigma_i}\Big)^2}},
 \end{equation}
 
\begin{equation}
    Accuracy = \frac{\text{RMSE}}{\text{Mean fruit size}}.
\end{equation}

Tables 1, 2 and 3 shows the RMSE and accuracy of using individual cameras. From the three cameras, the bottom camera gave the worst result with a height accuracy of 0.7473 and width accuracy of 0.8722. The middle camera gave the best height accuracy, with a result of 0.9162 and the top camera gave the best width accuracy of 0.9604. In terms of fill ratios, the top camera had a mean fill ratio of 0.940408, the middle camera had a mean fill ratio of 0.92659 and the bottom camera had a mean fill ratio of 0.813809. The bottom camera had the lowest mean fill ratio and the worst accuracy out of all three cameras, which suggests that it had the most occlusion, resulting in the measurement inaccuracy.
\begin{table}[h]
\caption{\label{tab:table-1} Measurement RMSE and accuracy using only the top camera.}
\begin{center}
\begin{tabular}{lllllllll}
\cline{1-3}
\multicolumn{1}{|l|}{Top Camera} & \multicolumn{1}{l|}{RMSE (mm)} & \multicolumn{1}{l|}{Accuracy  } &  &  \\ \cline{1-3}
\multicolumn{1}{|l|}{Height} & \multicolumn{1}{l|}{3.8889} & \multicolumn{1}{l|}{0.9013} &  &  \\ \cline{1-3}
\multicolumn{1}{|l|}{Width}  & \multicolumn{1}{l|}{1.8499} & \multicolumn{1}{l|}{0.9604} &  &  \\ \cline{1-3}
\end{tabular}
\end{center}

\caption{\label{tab:table-2} Measurement RMSE and accuracy using only the middle camera.}
\begin{center}
\begin{tabular}{lllll}
\cline{1-3}
\multicolumn{1}{|l|}{Middle Camera} & \multicolumn{1}{l|}{RMSE (mm)} & \multicolumn{1}{l|}{Accuracy} &  &  \\ \cline{1-3}
\multicolumn{1}{|l|}{Height} & \multicolumn{1}{l|}{3.3021} & \multicolumn{1}{l|}{0.9162} &  &  \\ \cline{1-3}
\multicolumn{1}{|l|}{Width}  & \multicolumn{1}{l|}{2.6819} & \multicolumn{1}{l|}{0.9426} &  &  \\ \cline{1-3}
\end{tabular}
\end{center}
\caption{\label{tab:table-3} Measurement RMSE and accuracy using only the bottom camera.}
\begin{center}
\begin{tabular}{lllll}
\cline{1-3}
\multicolumn{1}{|l|}{Bottom Camera} & \multicolumn{1}{l|}{RMSE (mm)} & \multicolumn{1}{l|}{Accuracy  } &  &  \\ \cline{1-3}
\multicolumn{1}{|l|}{Height} & \multicolumn{1}{l|}{9.9547} & \multicolumn{1}{l|}{0.7473} &  &  \\ \cline{1-3}
\multicolumn{1}{|l|}{Width}  & \multicolumn{1}{l|}{5.9701} & \multicolumn{1}{l|}{0.8722} &  &  \\ \cline{1-3}
\end{tabular}
\end{center}
\caption{\label{tab:table-4} Measurement RMSE and accuracy using the 3 camera system and selecting based on fill ratio.}
\begin{center}
\begin{tabular}{lllll}
\cline{1-3}
\multicolumn{1}{|l|}{} & \multicolumn{1}{l|}{RMSE (mm)} & \multicolumn{1}{l|}{Accuracy} &  &  \\ \cline{1-3}
\multicolumn{1}{|l|}{Height} & \multicolumn{1}{l|}{3.4920} & \multicolumn{1}{l|}{0.9114} &  &  \\ \cline{1-3}
\multicolumn{1}{|l|}{Width}  & \multicolumn{1}{l|}{2.6000} & \multicolumn{1}{l|}{0.9443} &  &  \\ \cline{1-3}
\end{tabular}
\end{center}
\end{table}

\begin{figure}[!htbp]
      \centering
        {\includegraphics[width=3.3in, height=5.5cm]{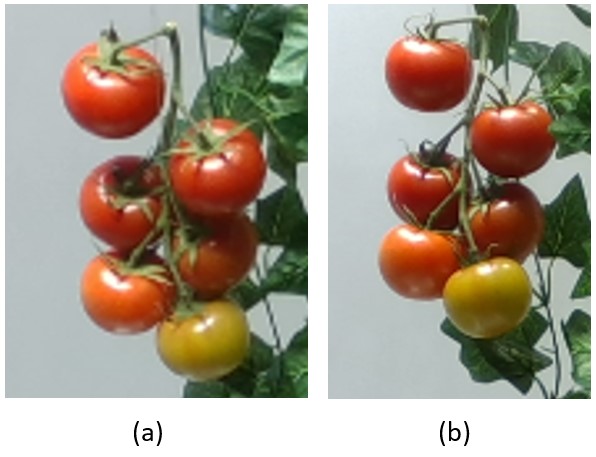}}
      \caption{(a) Tomato fruits viewed from the top camera, (b) Same fruits viewed from the middle camera.}
      \label{fig:view}
\end{figure}

With the 3 camera system implementation, the measurement with the highest fill ratio was selected, and it resulted in the height accuracy of 0.9114 and the width accuracy of 0.9443 as shown on Table 4. In the lab environment, the radius-based matching algorithm correctly matches all fruit detections, taking approximately 5 seconds for each photo. In general, the height accuracy was lower than the width accuracy, as the mean tomato height was smaller than the mean tomato width and the measurements would fluctuate largely depending on the pose of the fruit. As shown on Figure \ref{fig:view}, since the middle camera had the fruits viewing straight on, our method of measuring height from the top to bottom pixels of the bounding box was accurate. However, with the top camera, the fruits were slightly angled, causing an error with the height measurement.
\section{Conclusion}
To develop a tomato plant monitoring system, this paper has proposed a size measurement and visualization method using low cost RGBD cameras. This paper proposes a size measuring algorithm based on 3D distance calculation and occlusion estimation using circle fitting method. The algorithm was tested under a lab environment using 12 real tomato fruits on a bed of fake leaves to simulate occlusion. The results showed that the camera system and the algorithm was able to accurately measure the height and width of the tomatoes, giving a height accuracy of 0.9114 and the width accuracy of 0.9443. The mean tomato height was 39.4mm and the mean tomato width was 46.7mm for the 12 tomatoes, and the RMSE of height estimation was 3.492mm and RMSE of width estimation was 2.6mm.

Future work will focus on estimating pose of each tomato fruit and compensating the measurement to account for the pose of each fruit. The error in height measurement of tomatoes was larger than that of the width, mostly due to the height of tomatoes being shorter than the width and being more affected by the position of the fruit. Accounting for the pose of tomato fruit using various pose estimation methods such as calyx detection would be able to increase the accuracy of the measurement.

\section*{Acknowledgements}
    This research was funded by the New Zealand Ministry for Business, Innovation and Employment (MBIE) on contract UOAX2116, Artificial Intelligence-based Smart Farming System. Ho Seok Ahn is the corresponding author.

\end{document}